# A Model of Virtual Carrier Immigration in Digital Images for Region Segmentation

XIAODONG ZHUANG[1,2] and N. E. MASTORAKIS[1]

1. Technical University of Sofia, Industrial Engineering Department, Kliment Ohridski 8, Sofia, 1000 BULGARIA (mastor@wseas.org, http://www.wseas.org/mastorakis)
2. Qingdao University, Automation Engineering College, Qingdao, 266071 CHINA (xzhuang@worldses.org, http://research-xzh.cwsurf.de/)

*Abstract:* - A novel model for image segmentation is proposed, which is inspired by the carrier immigration mechanism in physical P-N junction. The carrier diffusing and drifting are simulated in the proposed model, which imitates the physical self-balancing mechanism in P-N junction. The effect of virtual carrier immigration in digital images is analyzed and studied by experiments on test images and real world images. The sign distribution of net carrier at the model's balance state is exploited for region segmentation. The experimental results for both test images and real-world images demonstrate self-adaptive and meaningful gathering of pixels to suitable regions, which prove the effectiveness of the proposed method for image region segmentation.

*Key-Words:* -Region segmentation, image analysis, virtual carrier immigration, self balancing, P-N junction

## 1 Introduction

In physics, the formation of P-N junction is fundamental for semiconductor devices [1,2]. The charge carriers in P-type and N-type semiconductor are holes and electrons respectively. The density of hole is high in P-type semiconductor, while electron has a high density in N-type semiconductor. When the materials of the two are put together with compact contact, diffusion of carries will happen at the interface of contact due to their density difference (i.e. carrier moving from high-density side to low density side). Meanwhile, a space charge region is established due to the carrier diffusion and recombination. The space charge region will grow with more diffusing of carrier, but it in turn causes the drifting of carriers which is at the opposite direction of diffusing. A demonstration of the physical P-N junction is shown in Fig. 1. The above dynamic process will reach a balance state when the drifting and diffusing of carriers get balanced, and a stable P-N junction will be established under that dynamic equilibrium [1,2].

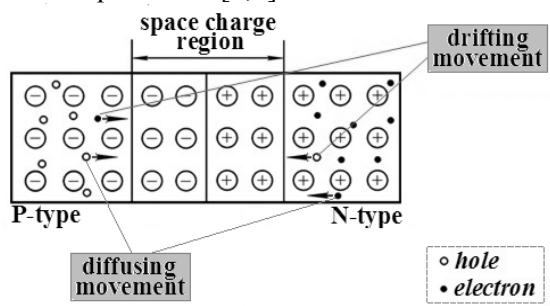

Fig. 1  The physical P-N junction

The region segmentation of image is a fundamental problem in image processing, which has significant value in both theoretic and practical research [3-9]. The basis of differentiation of two adjacent regions is their difference of image characteristics, such as grayscale, color, texture, etc. It is still an on-going and open research topic to segment image regions for various practical purposes [3-9]. In recent years, nature-inspired methods have attracted more and more research attention, in which the mechanisms in nature are imitated and adjusted in novel algorithms for image segmentation, and promising results have been obtained in such preliminary works [10-13]. In this paper, inspired by the physical P-N junction, the self-balancing mechanism of carrier diffusing and drifting is adopted in a novel segmentation framework, in which the region structure of image is formed by dynamic carrier diffusing and drifting.

In the following proposed model for region segmentation, the immigration of carrier caused by carrier density difference is called "diffusing", and that caused by virtual electric field is called "drifting". Of course, the formation of P-N junction is a microscopic physical process, while region segmentation is based on some algorithm implemented on computers. The algorithm proposed in this paper is inspired by the P-N junction, but not just a simulation of the physical process. In the following sections, the virtual carrier diffusing and drifting imitates the physical P-N junction, but the virtual electric field is defined artificially according to grayscale difference between adjacent image pixels. The proposed algorithm exploits the physical mechanism of carrier immigration, and introduces the factor of grayscale difference between the adjacent regions by the virtual electric field. Such difference between the algorithm and physical process should be noticed in reading the following sections.

## 2 The Model of Virtual Carrier Immigration in Digital Images

The proposed model for image segmentation is as follows. Suppose there are two categories of virtual carriers: positive and negative, which imitates the electron and hole. On the 2D plane of digital images, each pixel is modeled as a container of virtual carriers. Each pixel has four adjacent pixels (except those on image border), and correspondingly each carrier container has four adjacent containers. There is an interface between each pair of adjacent containers. Therefore, each pixel in the image corresponds to a carrier container with four different interfaces between its adjacent containers.

There are two features of the interface mentioned above. Firstly, the interface has permeability, which means the carriers at both sides of the surface can diffuse through it due to density difference. Secondly, there is a virtual electric field imposed on it, whose direction and intensity are determined by the grayscale difference between the corresponding two pixels connected by that interface. The virtual electric field is defined as:

$$e = K \cdot (g - g_a) \qquad (1)$$

where $e$ is the intensity of virtual electric field at the interface, $K$ is a predefined positive coefficient, $g$ is the grayscale of the pixel of interest and $g_a$ is that of its adjacent pixel. The direction of the electric field is defined according to the grayscale relationship of the pixel pair: the side with higher grayscale value has the higher electric potential. In another word, the container corresponds to higher grayscale value will attract negative carriers, while that corresponds to lower grayscale attracts positive carriers.

Moreover, the effect of each virtual electric field is limited to its corresponding interface only, and does not influence other interfaces. The virtual electric field on each interface plays a key role in the formation of diffusing-drifting balance of virtual carriers in the model. In such a way, the model is established consisting of virtual containers of carrier, the interface between adjacent containers, the virtual electric field, and also the virtual carriers. As mentioned before, it is supposed that there are two kinds of carriers in the system: positive and negative.

The evolution of the system is analyzed as follows. Initially, suppose the positive and negative carriers are of the same quantity, and each container has the same amount of carriers. Also suppose all the containers have the same volume, so that the density of carrier in a container is proportional to the amount of carrier in it. Therefore, there is no density difference of carriers between adjacent containers at that time, and there is no carrier diffusion at the beginning. However, with the effect of the virtual electric field at each interface, the carriers drift across the interfaces due to the virtual force applied by the electric field. The drifting then causes carrier density difference between two sides of the interface, which in turn makes the carriers to diffuse due to that density difference. Obviously, the diffusion has the opposite effect of drifting. Such dynamic process evolves until a balance between drifting and diffusion is reached. The proposed model is shown in Fig. 2 and Fig. 3. Fig. 2 shows the details of carrier immigration between two adjacent containers, while Fig. 3 shows the overall structure of the model on digital image. The balance state is worth of study for image segmentation, and the net carrier (i.e. the resultant amount after the offset of positive and negative carriers) in each container is of value in further analysis.

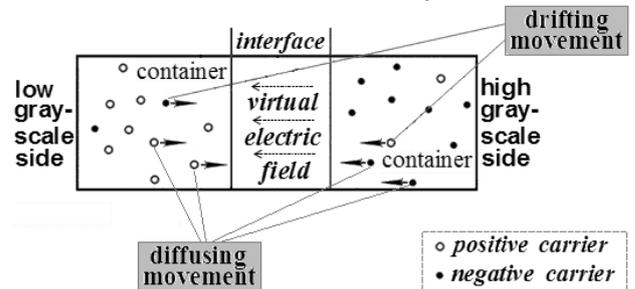

Fig. 2 Two adjacent containers in the proposed model

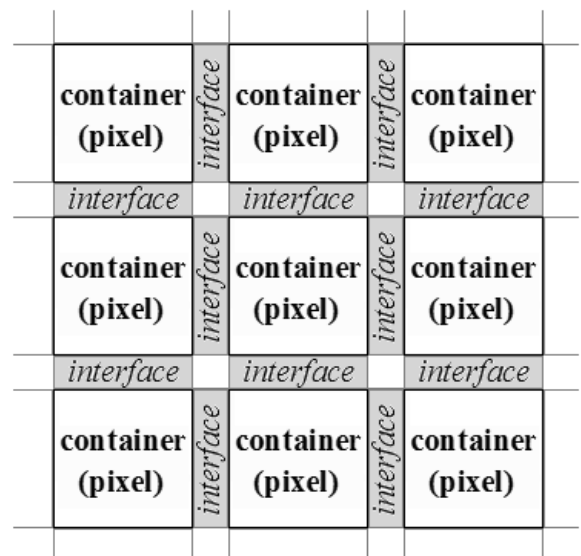

Fig. 3 The structure of the proposed model upon digital image

# 3 The Analysis of Virtual Carrier Immigration

The effect of the above dynamic process can be analyzed for two typical cases. The first case is for the pair of adjacent pixels with large grayscale difference, such as pixels at different sides of the region border. According to Equation (1), because the intensity of the virtual electric field is defined as proportional to the grayscale difference, in this case the virtual electric field is strong. The strong field will cause the carriers to drift, which quickly causes obvious density difference of carriers between the two sides of the interface. And the density difference in turn causes the opposite movement - the diffusion of carrier. Therefore, in this case the drifting of carriers caused by the virtual electric field is the dominant factor of carrier movement, which increases the carrier density difference between the two adjacent containers at the region border. On the other hand, the diffusing of carrier is an induced one, which functions as a factor to balance the drifting. In a word, for the case of large grayscale difference between adjacent pixels, the difference of the carrier density in their corresponding containers has an increasing tendency. Since there are two kinds of carriers, and their movements under the virtual electric field are opposite, in the above case the net carrier (i.e. the net charge) in one container will become the opposite to the other. That is to say, in this case the net charge in one container will become positive, while that in the adjacent one will become negative.

The second case is for the pair of adjacent pixels with small grayscale difference, such as adjacent pixels within the same region. In this case, the virtual electric field at the interface is weak because of the small grayscale difference. Therefore, in this case the dominant factor is the diffusing of carrier due to carrier density difference. Because the dominant diffusing process will decrease the difference of carrier density, there is a tendency to an identical distribution of carrier within a region. As a result, the containers within the same region tend to have the same sign of net carrier (i.e. all positive or all negative).

The obvious difference between the above two cases just satisfies the requirement of segmenting adjacent regions. With the immigration of carriers going on, at the boundary of two different regions, the carrier density difference is large between the two sides of the region boundary. On the other hand, the carrier density of the containers within a region has a tendency to an identical distribution by the diffusing process. The final result at balancing will reflect the differentiation of adjacent regions, which may provide useful clues for image segmentation. The simulation results in following sections provide experimental support to the above analysis.

A brief demonstration of the above dynamic process of virtual carrier immigration is shown in Fig. 4. Fig. 4(a) shows the initial state, in which all the containers have the same carrier density, and only drifting exists at the borders of two different grayscale region due to virtual electric field caused by grayscale difference (i.e. no diffusing of carriers at beginning). The later situation is shown in Fig. 4(b), where diffusing emerges between adjacent containers due to carrier density difference. The ideal final state in balance is shown in Fig. 4 (c), where the balance of diffusing and drifting is established at region borders, while the carrier density distribution becomes identical for the containers inside a homogeneous region of uniform grayscale. It is indicated in Fig. 4(c) that the distribution of net carrier at balance state can be exploited for region segmentation because the adjacent regions have opposite sign of net carrier.

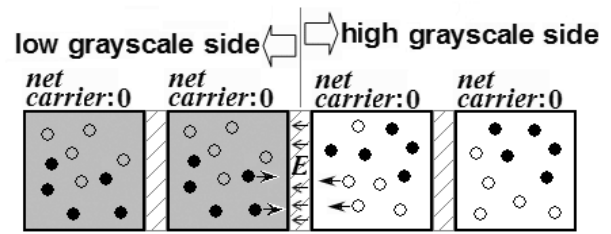

(a) the initial state

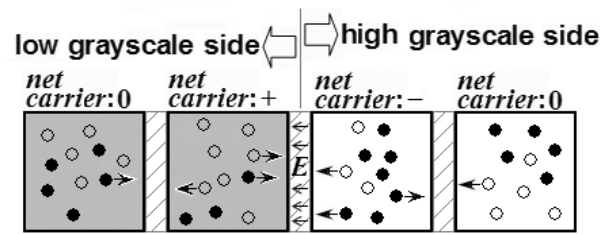

(b) the evolving progress

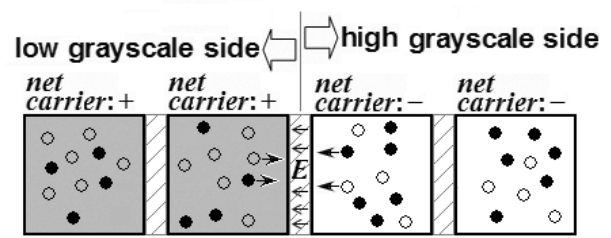

(c) the ideal finale state at balance

Fig. 4 A demonstration of the dynamic process of carrier immigration

Although the virtual electric field and the diffusing-drifting of carriers are defined locally at each interface between adjacent containers, the locally "connected" containers form an integrated whole. With the immigration of carriers from one container to another, the interaction between local adjacent pixels can affect other image areas gradually by the evolving of the system. Finally, balance may be reached at both local and global level, and some pattern of carrier distribution can emerge globally, which may be meaningful and useful to image segmentation. This is a unique nature of the proposed model compared with traditional local operators such as local edge detectors.

There is difference between the proposed model and the physical P-N junction. In physics, the basic factor is the carrier density difference between the P-type and N-type semiconductors. The carrier density difference causes the diffusion of carrier, which forms the electric field (i.e. the space charge region) at the interface. The electric field in turn induces the drifting of the carrier. However, the process in the proposed model for region segmentation is just different. For each pair of adjacent pixel, the virtual electric field is firstly defined according to the grayscale difference between adjacent pixels, and keeps unchanged in the whole evolving process. The virtual electric field causes the drifting of carriers, which causes carrier density difference between adjacent pixels (virtual containers). Then the density difference in turn makes the carriers to diffuse, which is an opposite movement to drifting. Therefore, the proposed model adopts the drifting-diffusing mechanism of carrier in physics, but it is not just only a simulation of physical P-N junction. Since the virtual electric field is the activating factor of the evolving process in the proposed model, the result at balance state after evolving will reflect pixel difference locally and also region difference globally.

# 4 Image Segmentation Based on Virtual Carrier Immigration

## 4.1 Model implementation by computer simulation

In the simulation of the model on computer, the simulation must be implemented in discrete steps (i.e. iteration by iteration). In one simulation step, the drifting speed of carrier (i.e. the amount of carrier immigrating from one container to the other in one iteration of simulation, or one simulation step) is defined directly proportional to the intensity of virtual electric field:

$$\Delta c_{drifting} = K_1' \cdot e \quad (2)$$

where $\Delta c_{drifting}$ is the amount of carrier drifting from one container into the other, $K_1'$ is a predefined positive coefficient, $e$ is the intensity of virtual electric field at the interface. According to Equation (1), $\Delta c_{drifting}$ is also proportional to the grayscale difference between the adjacent pixels:

$$\Delta c_{drifting} = K_1 \cdot (g - g_a) \quad (3)$$

where $\Delta c_{drifting}$ is the amount of carrier drifting from one container into the other, $K_1$ is a predefined positive coefficient, $g$ is the grayscale of the pixel of interest and $g_a$ is that of its adjacent pixel.

On the other hand, in one simulation step, the speed of diffusion has proportionality relationship with carrier density difference between the adjacent pixels. Suppose each container has the same size (or volume). Then the carrier density is proportional to the carrier amount in each container. Therefore, in the implementation the carrier density is substituted by carrier amount for diffusing:

$$\Delta c_{diffusing} = K_2 \cdot (c - c_a) \quad (4)$$

where $\Delta c_{diffusing}$ is the amount of carrier diffusing from one container into the other, $K_2$ is a predefined positive coefficient, $c$ is the net carrier amount in the container of interest and $c_a$ is that in its adjacent container.

For Equation (4), there is another issue to be discussed in the implementation. There are two types of carriers in the model: positive and negative. Each container has both types in it. In the evolving process, the two types of carrier immigrate by drifting and diffusing respectively. For each container, the net carrier is the combination of the two types after the offset between them. For simplicity in implementation, the immigration of carriers is measured by the amount of net carrier. In another word, the carrier density and the flow of carrier between containers are measured by net carrier amount.

The steps of implementing the model are as follows. At the beginning, the amounts of positive and negative carriers are equal in each container. Also suppose the amount is sufficient for arbitrary amount of carrier immigration in the simulation. Then the virtual electric field is calculated at each interface between adjacent containers, which is proportional to the grayscale difference between corresponding adjacent pixels. The detailed simulation step is as follows.

*Step*1 For each of the four interfaces of every virtual container (or pixel), do the following:

calculate the drifting amount of carrier due to virtual electric field; calculate the diffusing amount of carrier due to carrier density difference; sum the above two for all the 4 interfaces of a container to get its total change of net carrier amount; update the net carrier amount in that container;

*Step*2 After all the containers update their net carrier amount, calculate the average change of net carrier for all the containers. If the average change of net carrier is smaller than a predefined threshold, it is close enough to the balance state, and the simulation stops; otherwise, return to *Step*1 to begin a new iteration of simulation.

### 4.2 Simulation results for simple test images

Some of the experimental results for a series of test images are shown from Fig. 5 to Fig. 8. The simulation is implemented by programming in C.

Fig. 5 shows the results for the mostly simple case: the left and right half of the image are of different grayscale, which can specifically reveal the effect of the model on region borders. In order to demonstrate the process of carrier immigration step by step, the intermediate results at several specific simulation step numbers are recorded together with the final result. In the following results of net carrier sign distribution such as Fig. 5(b) to Fig. 5(f), the white and black points represent the positive and negative sign of net carrier respectively, and the gray points represent that the net carrier is zero (i.e. not a definite sign yet). In Fig. 5(b), it can be seen that shortly after the simulation starts, only the points close to the region border have a definite sign of net carrier. From Fig. 5(c) to Fig. 5(f), it is clear that the part of definite sign of net carrier expands with increasing simulation time. Finally, in Fig. 5(f) each point has a definite sign (positive or negative) of net carrier, and the final result of sign distribution just corresponds to the segmentation of the two halves of the image.

Fig. 6(a) shows another test image with a rectangle region. The intermediate results are shown from Fig. 6(b) to Fig. 6(f). It is clear that the positive sign part expands inward, while the negative sign part expands outward from the rectangle border with increasing simulation time. Similarly, the final result of sign distribution of net carrier in Fig. 6(g) can also provide a definite segmentation of the Test image 2.

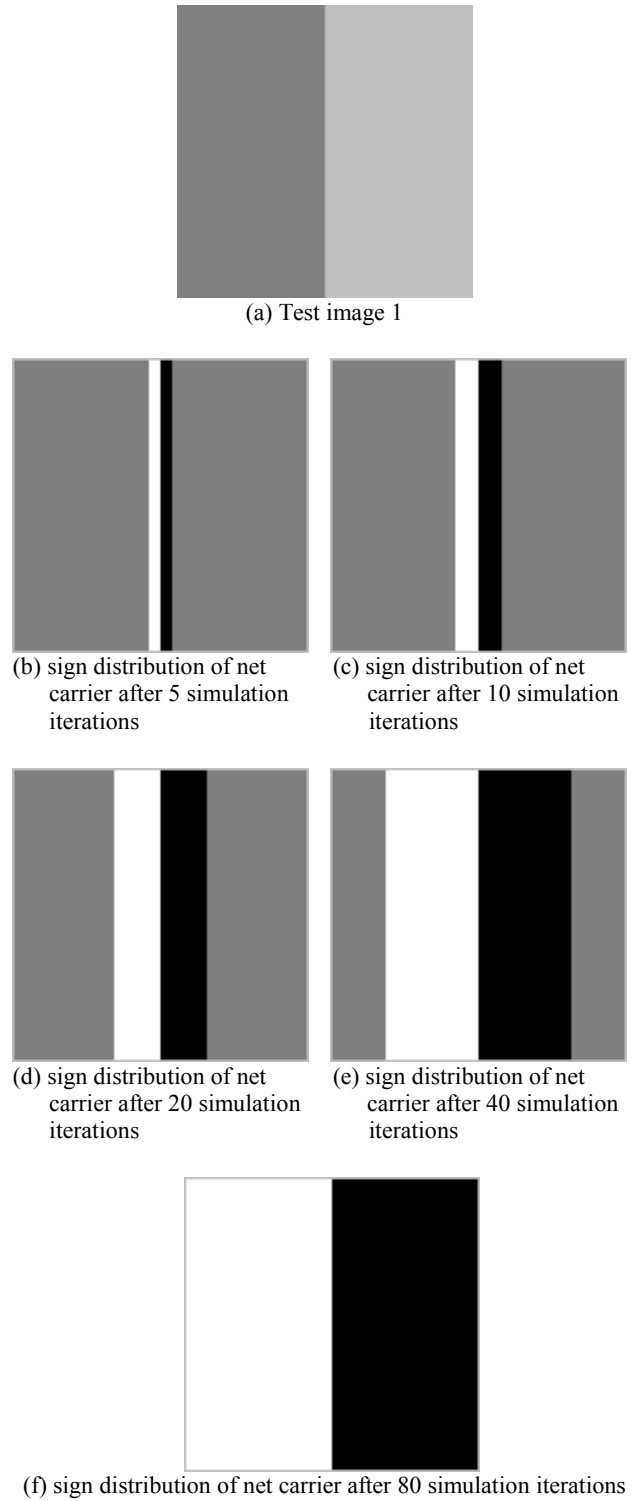

Fig. 5 The simulation results for Test image 1

(a) Test image 1
(b) sign distribution of net carrier after 5 simulation iterations
(c) sign distribution of net carrier after 10 simulation iterations
(d) sign distribution of net carrier after 20 simulation iterations
(e) sign distribution of net carrier after 40 simulation iterations
(f) sign distribution of net carrier after 80 simulation iterations

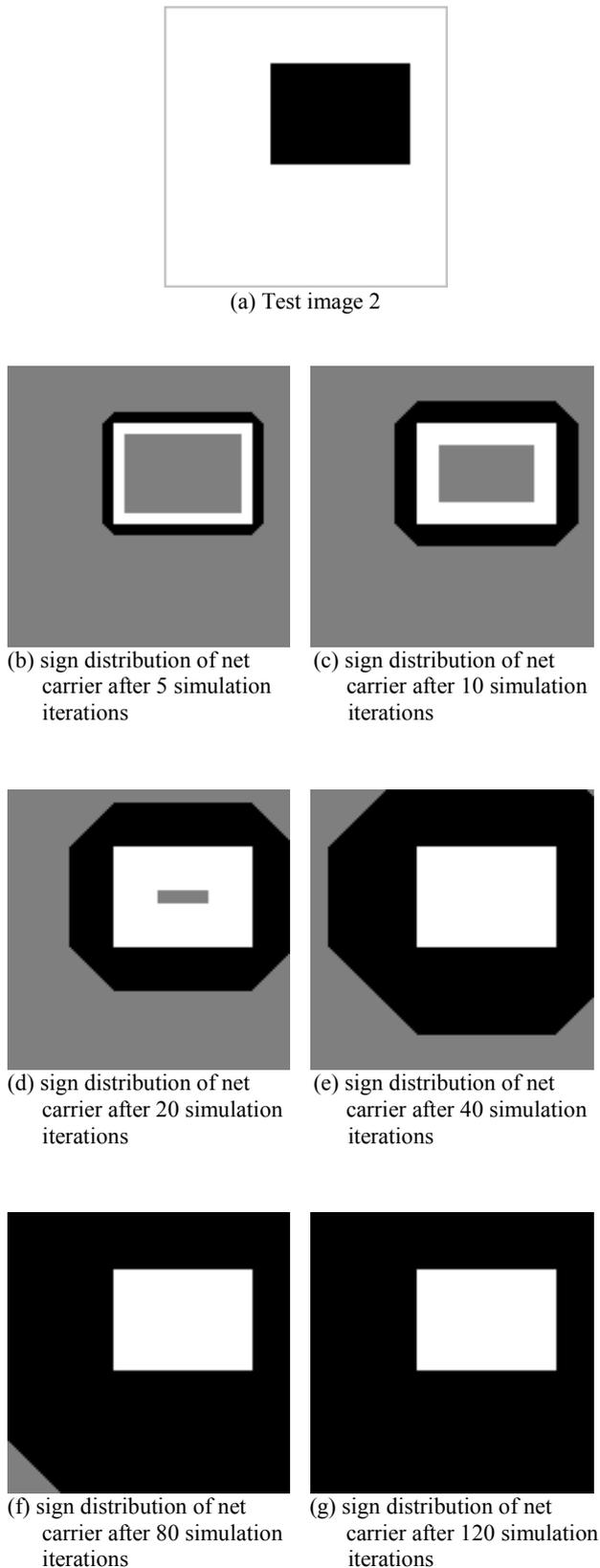

Fig. 6 The simulation results for Test image 2

Fig. 7(a) shows a test image with three regions of different shape and grayscale. The intermediate results are shown from Fig. 7(b) to Fig. 7(e). The process of carrier immigration step by step can be clearly seen. The final result in Fig. 7(f) almost implements the segmentation of regions, in which the three regions in Test image 3 have positive sign of net carrier, while the background has negative sign of net carrier.

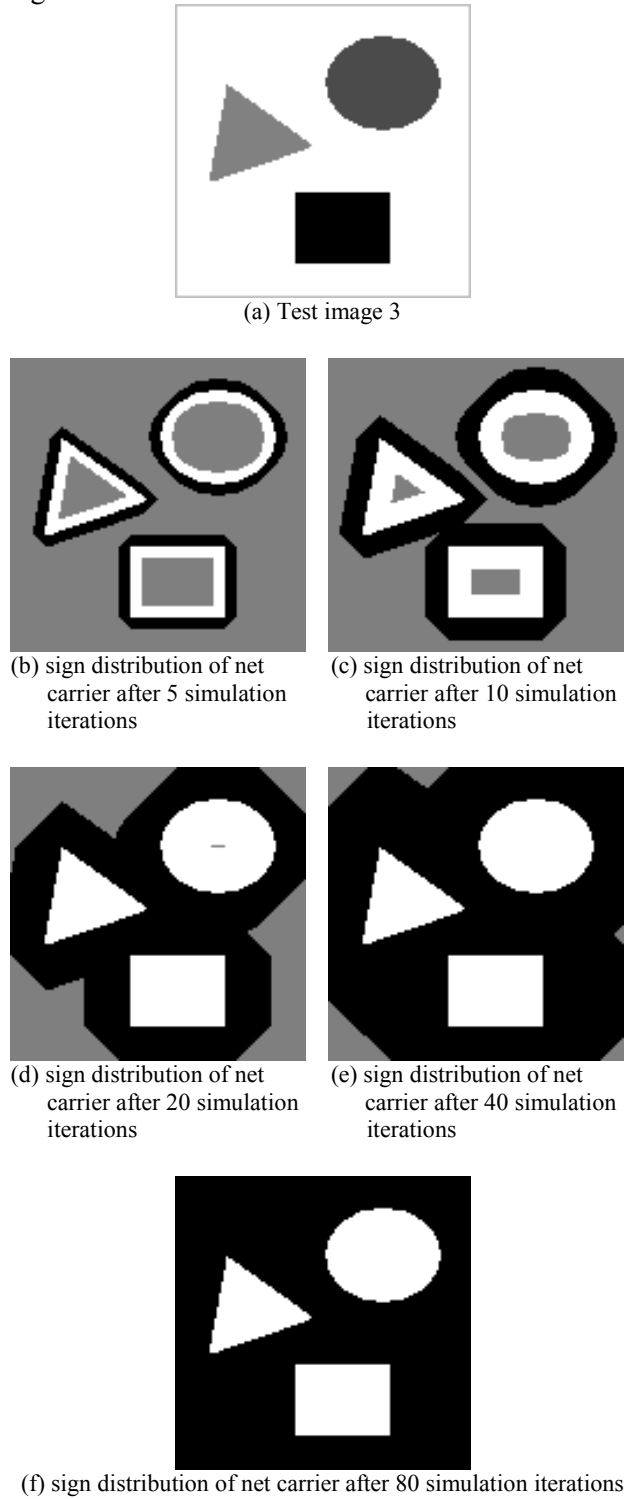

Fig. 7 The simulation results for Test image 3

The above interesting results for test images can be analyzed as follows. At region borders with large grayscale variation, the dominant factor of carrier immigration is the drifting caused by the strong virtual electric field. As the overall effect, for a pair of containers, the positive carriers tend to gather into one container, while the negative carriers gather into the other. If the electric field is defined as proportional to the grayscale difference, the container corresponding to higher grayscale pixel has the higher electric potential, which attracts negative carriers to gather. On the other hand, the positive carriers are attracted into the other container corresponding to lower grayscale. Such effect will increase the difference of net carrier amount between the adjacent containers, and the sign of the net carriers in them also tend to become opposite.

On the other hand, for a local area within a region, the grayscale difference is small. Correspondingly, the virtual electric field is also small. Here the dominant factor for carrier immigration is the density difference of carriers between adjacent containers, which makes the carriers diffusing, and produces a tendency of identical carrier distribution inside a region. As the result, the amount and sign of net carrier for the containers inside a region tend to become homogeneous.

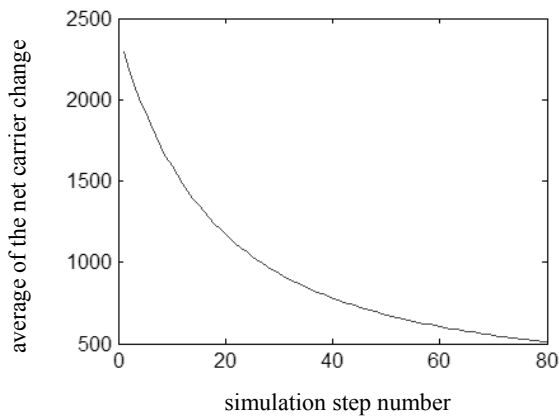

Fig. 8 The relationship between the average of the change of net carrier and the simulation time (for Test image 3)

To investigate the carrier immigration process quantitatively, the average of the absolute value of net carrier change for all the containers is calculated and recorded as a measurement of the convergence degree to the balance state. Fig. 8 shows the relationship between that average value and the simulation step number in the experiment for Test image 3. In Fig. 8, the average variation of net carrier decreases with the increasing of simulation time, which indicates the system approaches the balance state with the simulation going on.

Therefore, the evolving of the model can be summarized briefly: the net carrier difference across the region boundary is increased by carrier drifting caused by the strong virtual electric field there, which makes the sign of net carrier opposite between adjacent regions; then a specific sign of net carrier for a region spreads within that region by carrier diffusing. Therefore, different adjacent regions may be differentiated globally. This just suits the requirement of region segmentation. The global distribution of the sign of net carriers at balance state can provide effective basis for image segmentation.

### 4.3 Image segmentation based on the proposed model for real-world images

In the above experimental results for the test images, it is shown that the sign of net carrier are opposite in two adjacent regions, which can provide the basis of region division in images. In order to obtain the segmentation result from the sign distribution of the net carrier, a region grouping approach is proposed as following:

*Step***1**: Implement the simulation of the virtual carrier immigration as proposed in section 4.1;

*Step***2**: Obtain the sign distribution of the net carrier;

*Step***3**: Group the adjacent containers (i.e. image points) with the same sign of net carrier as connected points in same region. In the region grouping process, the adjacent pixels of the 4-connection (i.e. the upper, lower, left and right pixels) for an image point *p* is investigated. If any of the four adjacent containers (pixels) has the same sign of net carrier as *p*, it is grouped into the region which *p* belongs to. The obtained regions are the result of region segmentation for the image.

The obtained set of regions is the result of region segmentation. Fig. 9, Fig. 10 and Fig. 11 show the region segmentation results according to Fig. 5(f), Fig. 6(g) and Fig. 7(f), where different regions are represented by different gray-scale values.

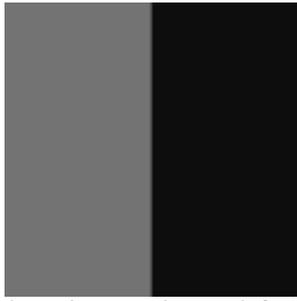

Fig. 9 The region grouping result for Fig. 5(f)

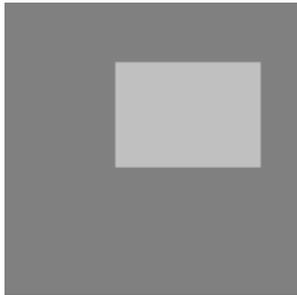

Fig. 10 The region grouping result for Fig. 6(g)

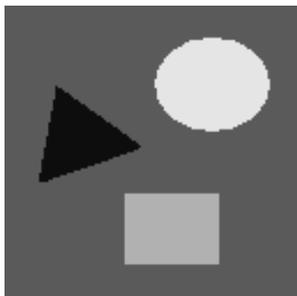

Fig. 11 The region grouping result for Fig. 7(f)

However, real world images consist of more complex regions than the simple test images. Moreover, gradual grayscale changes are commonly seen in natural images rather than sharp grayscale change at the region borders. To investigate the effect of the carrier immigration method on real world images, experiments are carried out for a series of real world images. For demonstration, some of the results are shown in Fig. 12 to Fig. 17, which are for the broadcaster image, the house image, the locomotive image, the peppers image and the medical heart image. The experimental results indicate that the proposed method can obtain large amount of regions (more than a hundred) because of the complexity of real world images. There are 498 regions obtained for the broadcaster image, 537 for the house image, 395 for the locomotive image, 177 for the peppers image, and 533 for the medical heart image. These results are shown in Fig. 12(c), Fig. 13(c), Fig. 14(c), Fig. 15(c) and Fig. 16(c) respectively.

To obtain practically useful segmentation result, a region merging method is proposed for the above region segmentation results based on the gray-scale similarity of adjacent regions. First, an expected number of remaining regions after merging is given (usually by trail). Then the following steps are carried out to merge regions until the expected region number is reached:

*Step*1: For each region in the image, calculate its average gray-scale value.
*Step*2: Find the pair of neighboring regions with the least difference of the average gray-scale, and merge them into one region.
*Step*3: If current region number is larger than the expected region number, return to *Step*1; otherwise, end the merging process.

For each real world image, the figures show the original image, the sign distribution of net carrier at the balance state, the region segmentation results by grouping, and also the result of region merging. In the sign distribution of net carrier, the white points represent positive net carrier, and black points represent negative net carrier. In the region segmentation results and region merging results, different regions are represented by different grayscale values.

For the broadcaster image, the remained region number after merging is 20 in Fig. 12(d).

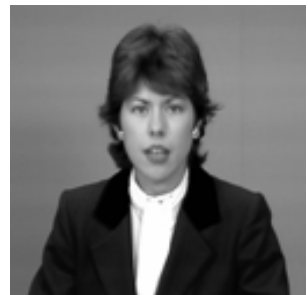 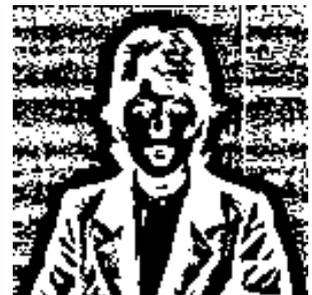

(a) the broadcaster image    (b) the sign distribution of $z$ coordinate at balance state

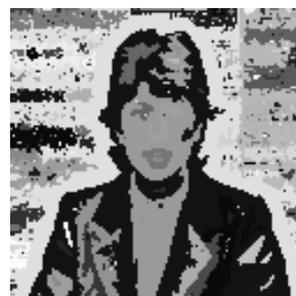 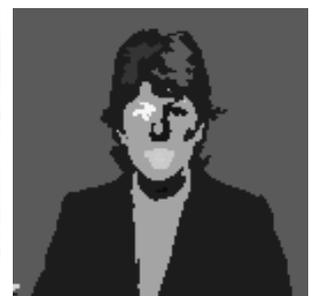

(c) the region segmentation result for (b)    (d) the merging result for (c)

Fig. 12 The experimental results for the broadcaster image

For the house image, the remained region number after merging is 50 in Fig. 13(d).

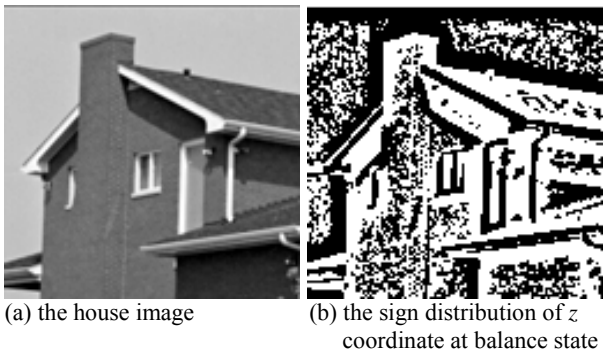

(a) the house image  (b) the sign distribution of $z$ coordinate at balance state

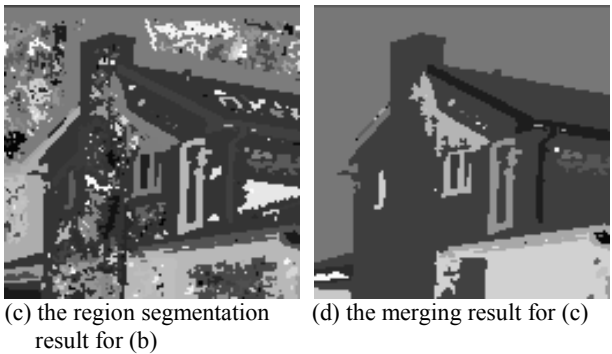

(c) the region segmentation result for (b)  (d) the merging result for (c)

Fig. 13 The experimental results for the house image

For the locomotive image, the remained region number after merging is 80 in Fig. 14(d).

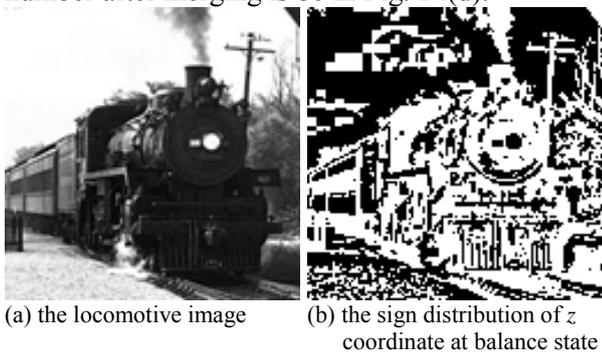

(a) the locomotive image  (b) the sign distribution of $z$ coordinate at balance state

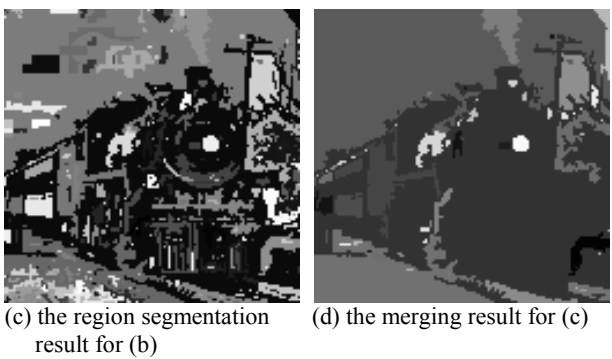

(c) the region segmentation result for (b)  (d) the merging result for (c)

Fig. 14 The experimental results for the locomotive image

For the peppers image, the remained region number after merging is 50 in Fig. 15(d).

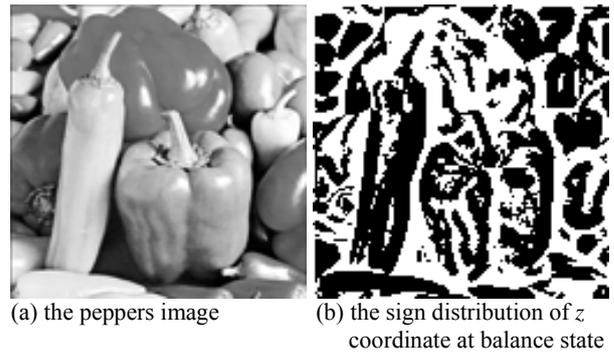

(a) the peppers image  (b) the sign distribution of $z$ coordinate at balance state

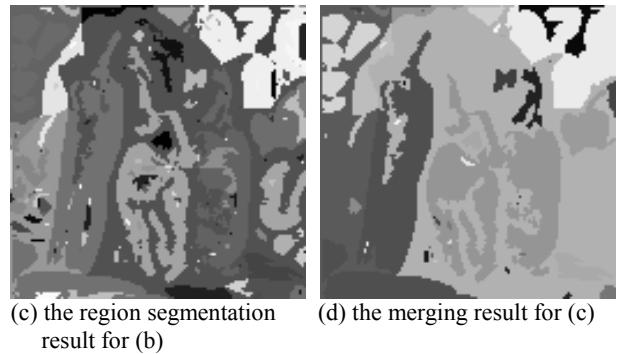

(c) the region segmentation result for (b)  (d) the merging result for (c)

Fig. 15 The experimental results for the peppers image

For the medical image of the heart, the remained region number after merging is 50 in Fig. 16(d). The merged result in Fig. 16(d) shows the heart structure clearly. Moreover, the average of the net carrier change for all the points is calculated and recorded as a measurement of the convergence degree to the balance state. Fig. 17 shows the relationship between that average value and the simulation time, which indicates that the process of carrier immigration approaches the balance state with the simulation going on.

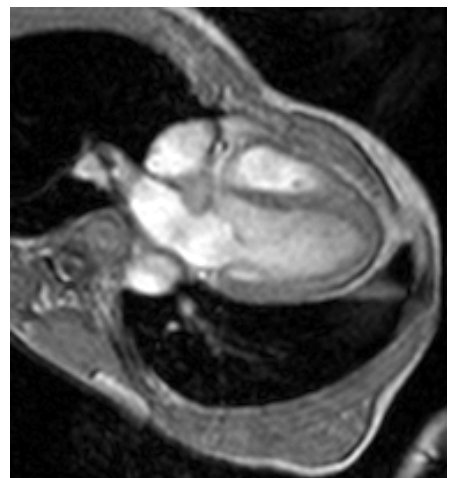

(a) the medical heart image

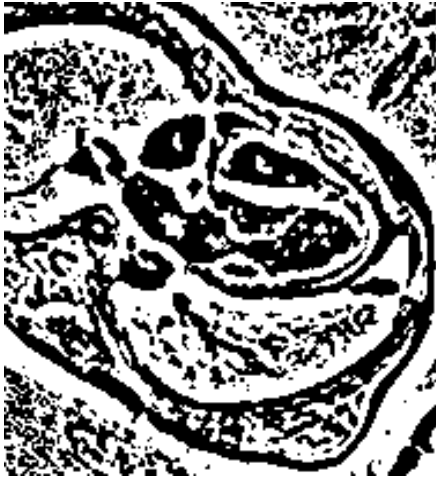

(b) the sign distribution of *z* coordinate at balance state

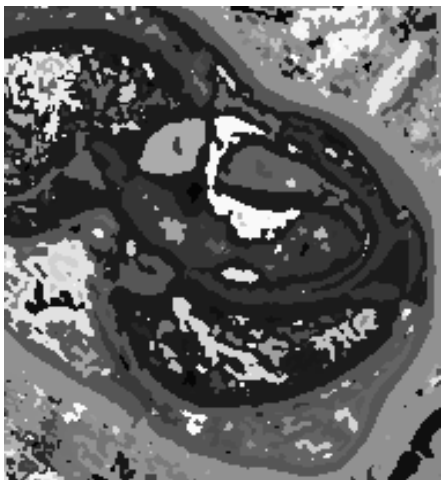

(c) the region segmentation for (b)

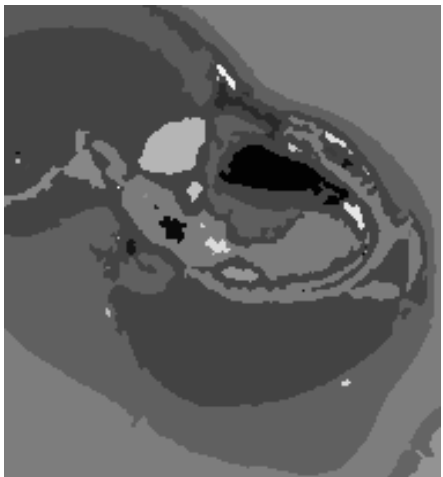

(d) the merging result for (c)

Fig. 16 The experimental results for the medical heart image

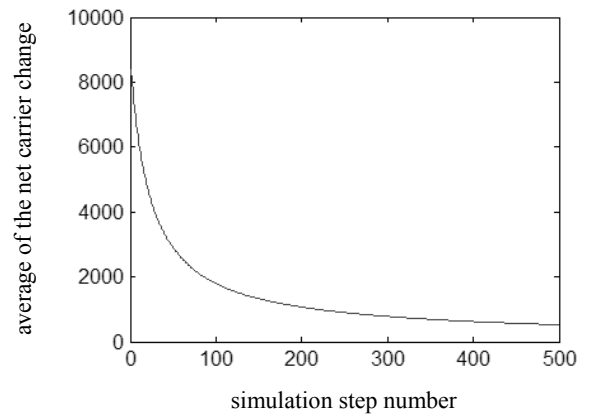

Fig. 17 The relationship between the average change of net carrier and the simulation time (for the medical heart image)

The above experimental results prove that the analysis of the model's dynamics in section 3 still holds for real world images. The proposed method is effective in segmentation of real world images. Relatively large amount of regions can be obtained by the grouping of the net carrier sign due to the complexity of real world images. From the results, it can be seen that main object regions can be well segmented, and some regions are segmented in perfect detail. However, at some part of object boundaries, two objects are not well separated due to reasons like grayscale similarity. It is indicated that other image features besides grayscale may be introduced into segmentation for improvements.

## 5 Conclusion and Discussion

In this paper, a new model of virtual carrier immigration on digital image is presented by imitating the diffusing and drifting of carriers in physical P-N junction. In the model, the virtual electric field between adjacent pixels is defined according to their grayscale difference, which is the major difference between the proposed model and real P-N junction. In the model, the two kinds of immigration are: carrier drifting caused by the virtual electric field, and the carrier diffusing caused by the carrier density difference. The direct local interaction and indirect global interaction of the above two carrier movements can lead to a balance state of carrier distribution, which provides clues for region segmentation. Image segmentation is implemented based on the sign distribution of net carrier at balance state, and a merging step is applied to get more comprehensible and useful segmentation results. The experimental results for test images and real world images prove the effectiveness of the proposed method.

As a novel method in the developing field of physics-inspired image processing technique, the proposed method has a strong physical background, which is a distinctive feature compared to those state of the art method. The mechanism underlying the evolving process of the method's simulation will be further studied, which may possibly be abstracted as a general methodology for region segmentation.

For improvement of the segmentation results, detailed properties of the proposed model will be studied in future work. Experiments will be carried out to investigate the segmentation results under various parameter configurations in the method. Color and texture feature will also be introduced into segmentation for possible improvement. And more reasonable merging process will also be explored as a necessary post-processing step to obtain better results of region merging.